\documentclass[twoside,11pt]{article}
\usepackage[accepted]{melba}
\usepackage{amsmath,amssymb,amsfonts}
 \usepackage{booktabs}
\usepackage{mathrsfs}
\usepackage{algorithmic}
\usepackage{graphicx}
\usepackage{textcomp}
\usepackage{url}
\usepackage{afterpage}
\usepackage{subcaption}
\usepackage{makecell}
\usepackage{tabularx}
\usepackage{hhline}
\usepackage{siunitx}
\usepackage{xr}
\usepackage{lscape}
\usepackage{float}
\usepackage{color,soul}
\usepackage[skip=0pt]{caption}
\def\BibTeX{{\rm B\kern-.05em{\sc i\kern-.025em b}\kern-.08em
    T\kern-.1667em\lower.7ex\hbox{E}\kern-.125emX}}

\newcommand{\I}{I}
\newcommand{\F}{F}
\newcommand{\Fhat}{\tilde{F}}
\newcommand{\Ihat}{\tilde{I}}
\newcommand{\kx}{k_x}
\newcommand{\ky}{k_y}

\newcommand{\af}{\sigma'}
\newcommand{\ai}{\sigma}
\newcommand{\un}{v_n}
\newcommand{\xn}{u_n}
\newcommand{\biasf}{b_n'}
\newcommand{\biasi}{b_n}
\newcommand{\ci}{\hat{u}_n}
\newcommand{\cf}{\hat{v}_n}
\newcommand{\wf}{w_n'}
\newcommand{\wi}{w_n}
\newcommand{\alphan}{\alpha_n}
\newcommand{\betan}{\beta_n}
\newcommand{\xnext}{u_{n+1}}
\newcommand{\unext}{v_{n+1}}
\newcommand{\sigm}{s}

\newcommand{\Joint}{\texttt{Interleaved}}
\newcommand{\Alternating}{\texttt{Alternating}}
\newcommand{\Image}{\texttt{Image}}
\newcommand{\Freq}{\texttt{Frequency}}

\newcommand\nw[1]{\textcolor{black}{#1}}

\melbaheading{2022:018}{https://www.melba-journal.org/papers/2022:018.html}{2022}{1-28}{10/2021}{06/2022}{Singh et al.}{}{}

\begin{document}
\title{Joint Frequency and Image Space Learning for MRI Reconstruction and Analysis}
\author{\name Nalini M. Singh \email nmsingh@mit.edu \\  
	\addr Computer Science and Artificial Intelligence Laboratory, MIT, Cambridge, MA, USA \\
	\addr Dept. of Health Sciences \& Technology, MIT, Cambridge, MA, USA
	\AND
	\name Juan Eugenio Iglesias \email e.iglesias@ucl.ac.uk \\
	\addr A. A. Martinos Center, Massachusetts General Hospital, Boston, MA, USA \\
	\addr Harvard Medical School, Cambridge, MA, USA \\
	\addr Centre for Medical Image Computing, UCL, London, UK \\
	\addr Computer Science and Artificial Intelligence Laboratory, MIT, Cambridge, MA, USA
	\AND
	\name Elfar Adalsteinsson \email elfar@mit.edu \\
	\addr Research Laboratory of Electronics, MIT, Cambridge, MA, USA \\
	\addr Dept. of Electrical Engineering \& Computer Science, MIT, Cambridge, MA, USA
	\AND
	\name Adrian V. Dalca \email adalca@mit.edu \\
	\addr A. A. Martinos Center, Massachusetts General Hospital, Boston, MA, USA \\
	\addr Harvard Medical School, Cambridge, MA, USA \\
	\addr Computer Science and Artificial Intelligence Laboratory, MIT, Cambridge, MA, USA
	\AND
	\name Polina Golland \email polina@csail.mit.edu \\
	\addr Computer Science and Artificial Intelligence Laboratory, MIT, Cambridge, MA, USA \\
	\addr Dept. of Electrical Engineering \& Computer Science, MIT, Cambridge, MA, USA
	\vskip-0.2in
}

\maketitle
\begin{abstract}
We propose neural network layers that explicitly combine frequency and image feature representations and show that they can be used as a versatile building block for reconstruction from frequency space data. Our work is motivated by the challenges arising in MRI acquisition where the signal is a corrupted Fourier transform of the desired image. The proposed joint learning schemes enable both correction of artifacts native to the frequency space and manipulation of image space representations to reconstruct coherent image structures at every layer of the network. This is in contrast to most current deep learning approaches for image reconstruction that treat frequency and image space features separately and often operate exclusively in one of the two spaces. We demonstrate the advantages of joint convolutional learning for a variety of tasks, including motion correction, denoising, reconstruction from undersampled acquisitions, and combined undersampling and motion correction on simulated and real world multicoil MRI data. The joint models produce consistently high quality output images across all tasks and datasets.  \nw{When integrated into a state of the art unrolled optimization network with physics-inspired data consistency constraints for undersampled reconstruction, the proposed architectures significantly improve the optimization landscape, which yields  an order of magnitude reduction of training time. This result suggests that joint representations are particularly well suited for MRI signals in deep learning networks.} Our code \nw{and pretrained models are} publicly available at \url{https://github.com/nalinimsingh/interlacer}.
\end{abstract}

\section{Introduction}
\label{sec:introduction}
Magnetic resonance imaging (MRI)~\citep{lauterbur1973image} acquires frequency space data and converts these measurements to images for visualization and downstream analysis. Practical imaging considerations often affect the data acquisition process. For example, motion occurs during acquisition~\citep{andre2015toward}, noise affects sensor readings~\citep{macovski1996noise}, and sub-Nyquist undersampling is routinely used to speed up data acquisition~\citep{lustig2008compressed}. Traditionally, the acquired frequency space signals are converted to image space reconstructions via an inverse Fourier transform, with each individual frequency space measurement contributing to all output pixels in the image space. As a result, local changes in the acquired frequency space data induce global effects on the entire output image. To produce accurate image reconstructions, modeling tools for Fourier imaging must correct these global artifacts in addition to performing fine-scale image space processing.

\begin{figure}
\begin{center}
\includegraphics[width=\linewidth]{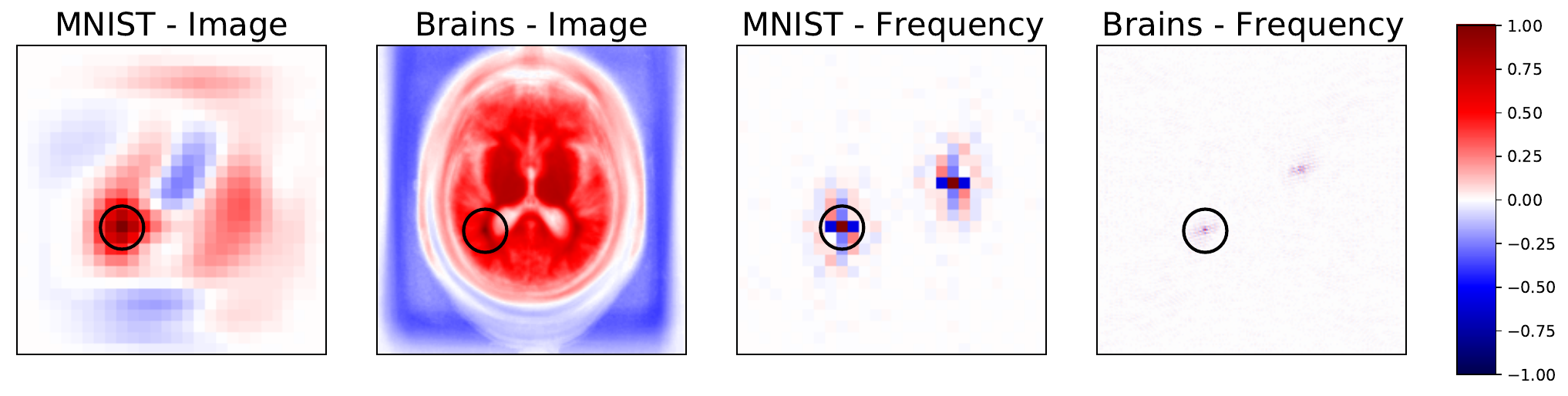}
\end{center}
\caption{Maps of correlation coefficients between a single pixel (center of circle) and all other pixels in image (left two panels) and frequency space (right two panels) representations of MNIST and a brain MRI dataset. All maps show strong local correlations useful for inferring missing or corrupted data in both spaces. Frequency space correlations also display conjugate symmetry characteristic of Fourier transforms of real images.}
\label{fig:correlations}
\end{figure}

\nw{Recently, neural networks have emerged as an alternative approach for MRI reconstruction~\citep{aggarwal2018modl,hammernik2018learning,hyun2018deep,lee2017deep,putzky2019invert,quan2018compressed,schlemper2017deep,sun2016deep,yang2017dagan,aggarwal2018modl, hammernik2018learning,cheng2018deepspirit,han2019k,zhu2018image,duffy2021retrospective, haskell2019network, johnson2019conditional, kustner2019retrospective, pawar2018moconet, shaw2020k, oksuz2019detection, usman2020retrospective,benou2017ensemble,jiang2018denoising,manjon2018mri}. Most existing architectures are based on purely frequency space representations or purely image space representations. Here, we propose and demonstrate joint frequency-image space representations that enable networks to learn a wide set of tasks including and beyond the extensively studied undersampled reconstruction.}
To motivate our approach, we examine the correlation structure for frequency and image space representations in Fig.~\ref{fig:correlations}. Local neighborhoods around a pixel exhibit strong correlations, suggesting that local convolution operations, which are widely successful on image space computer vision tasks, might also be useful when applied to frequency space data to capture this local structure. \nw{Convolutional operations in frequency space promise to enable direct correction of local frequency space artifacts corresponding to global image space effects, while convolutional image space processing facilitates complementary correction of artifacts that are best captured in the image domain.}

\subsection{Prior Work}
We study joint representations in the context of three corruption processes that arise during the imaging process.

\paragraph{Motion. } Previous retrospective motion correction strategies~\citep{batchelor2005matrix,haskell2018targeted} are cast as large, non-convex optimization problems with iterative solutions that are slow to compute. \nw{Deep learning methods~\citep{duffy2021retrospective, haskell2019network, johnson2019conditional, kustner2019retrospective, pawar2018moconet, shaw2020k, usman2020retrospective} solve the motion correction problem with a neural network operating purely in the image space, even though motion artifacts are induced directly in the frequency space during data acquisition. An alternative approach has been demonstrated recently that detects motion directly on frequency space data, followed by motion correction via an image space network~\citep{oksuz2019detection}.}

\paragraph{Noise. } Previous work on MRI denoising applies classical signal processing techniques including filtering~\citep{manjon2008mri} and wavelet-based methods~\citep{anand2010wavelet,nowak1999wavelet}. Deep learning methods employ convolutional networks solely on image space data~\citep{benou2017ensemble,jiang2018denoising,manjon2018mri}. 

\paragraph{Undersampling. } Classical undersampled reconstruction techniques either construct the output image as a least-squares estimate from the acquired frequency space data~\citep{pruessmann1999sense} or  combine convolutional filters in the frequency space with an inverse Fourier transform~\citep{griswold2002generalized,lustig2010spirit}.  Many deep learning methods apply convolutions to image space reconstructions of the acquired undersampled frequency data~\citep{aggarwal2018modl,hammernik2018learning,hyun2018deep,lee2017deep,putzky2019invert,quan2018compressed,schlemper2017deep,sun2016deep,yang2017dagan}. \nw{To improve the quality and fidelity of the reconstruction, the convolutional layers can be combined into an architecture that emulates unrolled optimization, with a convolutional regularizer coupled with a physics-inspired data consistency constraint that is enforced after each iteration~\citep{aggarwal2018modl, hammernik2018learning}.} Alternatively, the convolutional architectures can act directly on the frequency space data~\citep{akccakaya2019scan,cheng2018deepspirit,han2019k}. The notably different  AUTOMAP architecture uses fully-connected layers to  convert frequency space data to the image space and then applies further image space convolutions~\citep{zhu2018image}, incurring prohibitive memory complexity of $\mathcal{O}(N^4)$ for a~$N\times N$ image.

\nw{
\noindent More recently, solutions that combine frequency and image space convolutions have been demonstrated in the context of undersampled reconstruction. One approach is to combine separately trained pure frequency and pure image space networks into a common architecture~\citep{eo2018kiki, souza2019hybrid, wang2019accelerated}. The most closely related work to ours integrates frequency and image space blocks within the same network~\citep{zhou2020dudornet}, effectively implementing one of the two variants we consider in this paper. Here we propose an additional layer architecture that also tightly couples frequency and image space representations and evaluate both variants on a wide variety of tasks, well beyond the undersampled reconstruction scenario for which the previously combined architectures have been proposed. 
}

\nw{
In our experiments, a basic network that simply concatenates joint layers outperforms its pure frequency and image counterparts across a large set of artifacts and reconstruction quality metrics. To investigate how the joint layer architecture interacts with the data consistency constraints often used in undersampled reconstruction, we train the basic network with such a constraint and observe that it compares favorably with the state of the art task-specific undersampled reconstruction networks~\citep{eo2018kiki,schlemper2017deep} that also incorporate a data consistency constraint. Moreover, we probe the relationship between the proposed joint layers and the widely used unrolled optimization architectures by replacing image convolutional layers with our joint layers in a state of the art  unrolled optimization network, MoDL~\citep{aggarwal2018modl}. Using the proposed joint layers improves the training landscape and reduces training time by about an order of magnitude.
}

\nw{
To summarize, our contributions are as follows:
\begin{enumerate}
\item We define two task-independent convolutional layer architectures that tightly couple frequency and image representations of an input image that can be used in conjunction with unrolled optimization, data consistency constraints, and other sophisticated strategies for building and training reconstruction neural networks.
\item We demonstrate in simulation experiments that joint networks outperform pure image or pure frequency space networks for reconstructing high quality images in the presence of (i) extreme motion, (ii) heavy noise, and (iii) combination of artifacts, such as motion and undersampling. 
\item We demonstrate that the proposed joint learning strategy is compatible with a data consistency constraint and performs favorably relative to state-of-the-art networks specifically designed for the undersampled reconstruction task. 
\item We demonstrate on complex-valued, multicoil, real world data that incorporating joint layers into unrolled optimization networks results in more effective training and an order of magnitude decrease of training time, suggesting that the proposed architectures are particularly well suited for image representation in MRI reconstruction networks.
\end{enumerate}
}

 This paper is organized as follows. In the next section, we define the proposed layer and network architectures. Section 3 provides the  implementation details and describes our ablation studies. Section 4 reports experimental results, followed by the discussion of the proposed layers, their limitations, and  conclusions in Section 5.

\section{Joint Networks}
\label{sec:joint_networks}
MRI acquires Fourier transform measurements, referred to as k-space data. We assume a 2D multislice MRI acquisition. For each slice in this setup, the goal of image reconstruction is to generate an image~$\I$ from the acquired Fourier transform measurements~$F=\mathscr{F}\{I\}$. Classically, this reconstruction is computed via a 2D inverse Fourier transform, producing an estimated image~$\hat{I}=\mathscr{F}^{-1}\{F\}$. In practice, corrupted and possibly undersampled measurements~$\Fhat$ are acquired instead of~$\F$, and the goal is to estimate the desired image~$\I$ from the corrupted signal~$\Fhat$. Many strategies exist for selecting which measurements to acquire in frequency space. Here we consider Cartesian sampling, where measurement coordinates~$\kx$ and~$\ky$ are evenly sampled across the 2D Fourier plane, but our method can be generalized to other acquisition schemes. In this section, we define two neural network layer variants that combine image and frequency space convolutional features, referred to as \Joint~and \Alternating, specify the network architectures, and describe the learning procedure.

\subsection{Joint Layer Structures}
\begin{figure}[t]
    \begin{center}
    \includegraphics[width=\columnwidth]{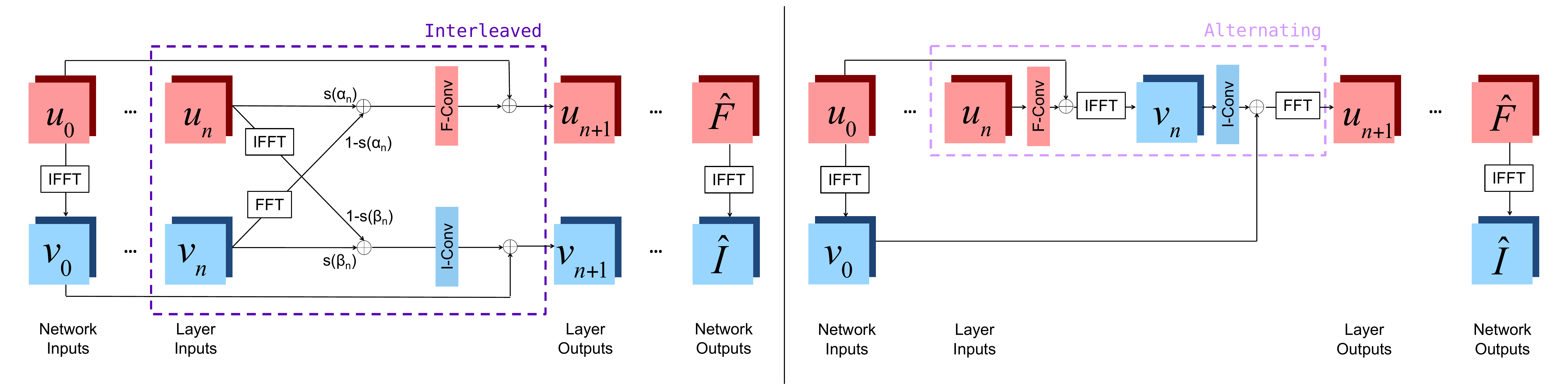}
    \end{center}
\caption{The \Joint~(left) and \Alternating~(right) layers, embedded within full network architectures. Each `F-Conv' or `I-Conv' block applies Batch Normalization (BN), a convolution, and an activation function in the frequency or image space, respectively.}
\label{fig:layer}
\end{figure}
Fig.~\ref{fig:layer} illustrates the layer structures of the two joint networks. We use~$\xn$ to denote the frequency space input and~$\un$ to denote the image space input of layer~$n$. Thus,~$u_0=\Fhat$ and~$v_0=\mathscr{F}^{-1}\left\{u_0\right\}$ represent the frequency space and image space inputs to the network. 

In the \Joint~setup, layer inputs are combined via \textit{learned}, layer-specific mixing parameters~$\alphan$ and~$\betan$ that parameterize the sigmoid function~$s(x)=(1+e^{-x})^{-1}$ to constrain the mixing coefficients to (0,1):
\begin{equation}
\label{eqn:layer_int}
\begin{alignedat}{5}
\ci &= \sigm(\alphan)\,\xn+(1-\sigm(\alphan))\,\mathscr{F}\left\{\un\right\}, \\
\cf &= \sigm(\betan)\,\un+ (1-\sigm(\betan))\,\mathscr{F}^{-1}\left\{\xn\right\}.
\end{alignedat}
\end{equation}
Real and imaginary parts of inputs are represented as separate channels at each layer and are joined appropriately to form complex numbers when computing the Fourier transform~$\mathscr{F}\left\{\cdot\right\}$ or its inverse. Next, the layer applies batch normalization (BN), a convolution, and an activation function with a skip connection to produce the outputs: 
\begin{equation}
\label{eqn:next_layer}
\begin{alignedat}{5}
\xnext &= \ai(\wi \circledast \textrm{BN}(\ci) + \biasi)+u_0, \\
\unext &= \af(\wf \circledast \textrm{BN}(\cf) + \biasf)+v_0,
\end{alignedat}
\end{equation}
where~$(\wi,\biasi)$ are learned frequency space convolution weights and biases,~$(\wf,\biasf)$ are learned image space convolution weights and biases, and~$\ai(\cdot)$ and~$\af(\cdot)$ are activation functions specific to the frequency space and image space network components, described later in this section. 

This layer architecture is a generalization of networks that operate purely in frequency space, obtained by choosing~$s(\alphan)=1$ and~$s(\betan)=0$, and of networks that operate purely in image space, that arise when~$s(\alphan)=0$ and~$s(\betan)=1$. When~$0<s(\alphan)<1$ and~$0<s(\betan)<1$, this layer represents a function that cannot be expressed solely via pure image or frequency space convolutional layers that do not invoke the Fourier transform or its inverse. \nw{Note that the frequency output $u_n$ of layer $n$ is not required to be the Fourier transform of the layer's image output $v_n$, only that the mixing is applied to either two frequency space outputs or two image space outputs. This additional flexibility ensures that $u_n$ and $v_n$ are not entirely redundant and the network learns the right features to capture MRI structure based on the input data and the task at hand.
}

In the \Alternating~setup, each layer sequentially incorporates frequency and image space convolutions with the appropriate batch normalization and activation function:
\begin{equation}
\label{eqn:alternating_next_layer}
\begin{alignedat}{5}
\un &= \mathscr{F}^{-1}\left\{\ai(\wi \circledast \textrm{BN}(\xn) + \biasi)+u_0\right\}, \\
\xnext &= \mathscr{F}\left\{\af(\wf \circledast \textrm{BN}(\un) + \biasf)+v_0\right\},
\end{alignedat}
\end{equation}
i.e., the reconstruction alternates between convolutions in the frequency and image space. A version of this architecture was previously introduced as part of a task-specific network for undersampled reconstruction~\citep{zhou2020dudornet}.

\nw{For both joint architectures, the frequency space convolutions represent element-wise multiplications in the image space. Since the convolution kernels have limited width, the learned convolutions cannot represent all such element-wise multiplications, but instead parameterize the subset whose 2D Fourier transform is zero outside of a central region. Coupled with nonlinearities in the frequency space, these operations enable the network to use global,  spatially varying operations not captured by image space convolutions.}

Although both of these layers explicitly include the Fourier transform and its inverse, no parameters are associated with those transforms. Thus, we learn only convolutional weights, biases, and possibly mixing coefficients. Since our networks incorporate Fourier transforms, they have an overall $\mathcal{O}(N^2 \log N)$ space complexity for~$N \times N$ images.

\subsection{Activation Functions} \label{section:Nonlinearity}
\nw{Adopting the standard practice of using the ReLU nonlinearity for image data, we define $\af(x) = \mbox{ReLU}(x)$ for all convolutions in the image space. This operation is applied separately to  real and imaginary channels of each image space convolution output (Trabelsi et al, 2018).}  However, the zero-gradient of this nonlinearity for negative values is ill-suited for networks that operate on frequency space data, as individual inputs can take on a large range of positive and negative values. We introduce an alternative nonlinear activation function that we apply to both the real and imaginary channels of each frequency space convolution output:
\begin{equation} 
    \label{eq:custom_nonlinearity}
    \ai(x) = x + ReLU\left(\frac{x-1}{2}\right)+ReLU\left(-\frac{x+1}{2}\right).
\end{equation}
This nonlinearity's magnitude increases with that of the input everywhere, while preserving the distinction between positive and negative inputs. We found that networks using this nonlinearity consistently outperformed networks that employed ReLU activation functions on frequency space convolution outputs.  

\subsection{Learning}
The networks evaluated in this paper can be trained with any differentiable loss function~$\mathcal{L}$. In our experiments, we investigate a wide variety of loss functions. We train the joint network~$f(\cdot;\theta_{\mathrm{f}},\theta_\mathrm{i})$ for image reconstruction by optimizing a set of frequency space parameters~$\theta_\mathrm{f}$ and a set of image space parameters~$\theta_\mathrm{i}$ over the training dataset~$\mathcal{D}=\{(\Fhat_{m},I_{m})\}$ using stochastic gradient descent-based strategies to obtain
\begin{equation}
\left(\theta_{\mathrm{f}}^*,\theta_{\mathrm{i}}^*\right)= \arg\min_{(\theta_{\mathrm{f}},\theta_{\mathrm{i}})}
\sum_{m=1}^{|\cal{D}|}\mathcal{L}\left (I_m,\mathscr{F}^{-1}\left(f(\Fhat_m;\theta_{\mathrm{f}},\theta_{\mathrm{i}})\right)\right ),
\end{equation}
where~$\theta_\mathrm{f}$ and~$\theta_\mathrm{i}$ depend on the setup of the joint layer.

\section{Implementation Details and Ablation Architectures}
\label{sec:implementation_details}
We construct each joint network to contain 10 joint frequency and image space layers. \nw{We performed a hyperparameter sweep and observed that the accuracy of reconstruction on the validation set  stopped improving for networks that included more than 10 joint layers.} A single 2D convolutional layer acts on the frequency space output $u_{10}$ of the final joint layer to produce the final 2-channel complex output~$\hat{F}$. The estimated image~$\hat{I}$ is the inverse Fourier transform of the network's output, i.e.,~$\hat{I}=\mathscr{F}^{-1}\{\hat{F}\}$. All convolution blocks within both types of joint layers have kernel size 3x3 and 64 output features, resulting in a total of 670,622 parameters for the \Joint~network and 706,438 parameters for the \Alternating~network.

To evaluate the utility of combined frequency and image space layers as a network building block for manipulating Fourier imaging data, we compare performance of the \Joint~and \Alternating~architectures to two similarly structured baseline architectures with only frequency or only image space operations.

First, we create an architecture \Freq~that performs convolutions only on frequency space data and train the network $g(\cdot;\theta_\mathrm{f})$ to identify frequency space parameters
\begin{equation}
\theta_{\mathrm{f}}^* = \arg\min_{\theta_{\mathrm{f}}}
\sum_{m=1}^{|\cal{D}|}\mathcal{L}\left (I_m,\mathscr{F}^{-1}\left(g(\Fhat_m;\theta_{\mathrm{f}})\right)\right ).
\end{equation} 
The network contains 20 convolution layer to match the joint networks' 10 pairs of 2 convolution layers. As in the \Joint\ and \Alternating~networks, each convolution layer has kernel size 3x3 and 64 output features, followed by the final, two-feature 2D convolutional layer, resulting in 706,438 parameters. \nw{This network captures the convolution strategy used in~\citep{akccakaya2019scan,han2019k,kim2019loraki}, which incorporate frequency space convolutions in the context of other task-specific architectures and loss choices.}

We also implement an image space network \Image. The network $g(\cdot;\theta_{\mathrm{i}})$ is trained by optimizing
\begin{equation}
\theta_{\mathrm{i}}^* = \arg\min_{\theta_{\mathrm{i}}}
\sum_{m=1}^{|\cal{D}|}\mathcal{L}\left (I_m,g\left(\mathscr{F}^{-1}\left(\Fhat_m\right);\theta_{\mathrm{i}}\right)\right).
\end{equation}
This network's architecture is identical to that of \Freq~and also contains 706,438 parameters, but it operates on image space data. \nw{This network captures the convolution strategy used in prior work that incorporates image space convolutions with task-specific architectures and loss function choices, e.g., unrolled optimization  and data consistency constraints~\citep{aggarwal2018modl,hammernik2018learning,haskell2019network,hyun2018deep,kustner2019retrospective,lee2017deep,manjon2018mri,pawar2018moconet,putzky2019invert,quan2018compressed,schlemper2017deep,sun2016deep,yang2017dagan}.}

We initialize all convolution weights using the He normal initializer~\citep{he2015delving} and use the Adam optimizer~\citep{kingma2014adam} (learning rate 0.001) until convergence. We initialize~$s(\alpha)$ and~$s(\beta)$ to 0.5.  Training each model requires one day on an NVIDIA RTX 2080 Ti GPU. Our code \nw{and pre-trained models} for each of these networks is available at \url{https://github.com/nalinimsingh/interlacer}.

\section{Experiments}
In this section, we evaluate the proposed joint layers in a set of experiments that progress from simulated data and basic networks to real world complex-valued multicoil MRI measurements and unrolled optimization frameworks with physics-inspired data consistency constraints. The experiments in this section are performed on brain MRIs from multiple datasets. Additional experiments on FastMRI single coil knee MRI, including comparisons with the top methods on FastMRI leaderboard, are provided in Appendix A.

\subsection{No Data Consistency}
\label{sec:exp1}
In this section, we present experiments where no data consistency contraint is employed in training our networks. These experiments directly compare the performance of the different layer types described in Sections~\ref{sec:joint_networks} and~\ref{sec:implementation_details}. These experiments are particularly useful for understanding the relative performance of these methods in settings where direct data consistency may not be desirable because the acquired data is corrupted by an artifact.

\paragraph{Data. } In this experiment, we simulate artifacts of interest in a set of 6,276 T$_1$-weighted brain MRI images from patients aged 55-90 collected as part of the Alzheimer's Disease Neuroimaging Initiative (ADNI)~\citep{mueller2005alzheimer}. We select the central 2D axial image of each volume for training and evaluation. To simulate acquired data, we apply the 2D Fourier transform to each image. After simulating the artifacts as described below, we normalize each input and output training pair by dividing by the maximum value in the corrupted image. \nw{The k-space data were zero-padded in this dataset during the original image reconstruction process, prior to our simulations. As a result, the quantitative results from these experiments do not represent model performance when deployed on raw, acquired k-space data~\citep{shimron2022implicit}. Instead, these experiments probe the relative performance of competing methods on tasks for which large datasets of raw k-space are not readily available, such as motion correction and denoising. Subsequent experiments with raw, acquired frequency space data that have not been padded demonstrate that the proposed joint layers can also handle non-padded data.
}
We split the dataset into 4,115 training images, 2,061 validation images, and 100 test images such that no subjects are shared across the training, validation, and test sets. Preliminary experiments and hyperparameters are evaluated on the validation dataset; the test set is only used for computing the performance statistics.

\paragraph{Training Loss and Evaluation Metric. }
\nw{We train \Freq, \Image, \Joint, and \Alternating~networks described in Section 3 using L1 loss on the real and imaginary components of the output and employ the SSIM scores~\citep{wang2004image} between the ground truth and reconstructed magnitude images to evaluate the quality of reconstruction on the test set.}

\subsubsection{Experimental Setup}
\label{section:ExperimentalSetup}

\paragraph{Motion.} \label{section:BrainMotion}
 Imaging subjects may move as measurements are being acquired at different points in the Fourier space. In practice, all points within a single line~$F(\cdot,\ky)$ in frequency space are acquired rapidly together. Thus, it is commonly assumed that no motion occurs during acquisition of a single frequency space line. \nw{In this work, we use a rigid-body motion model for motion that occurs between acquisitions of successive lines.} 

If the imaged subject is affected by a rotation~$\phi_{\ky}$ about the origin, a horizontal translation~$\Delta x_{\ky}$, and a vertical translation~$\Delta y_{\ky}$ during acquisition of line $\ky$, the acquired signal corresponds to the rigidly transformed image~$\Ihat_{\ky}$
\begin{align}
      \Fhat\left (\cdot,\ky\right ) = \mathscr{F}&\left\{ \Ihat_{\ky} \right\} \left (\cdot,\ky\right ), \quad \mathrm{where} \label{eq:motion2} \\
      \Ihat_{\ky}\left (x,y\right ) = I\bigl(&\left(x-\Delta x_{\ky}\right)\cos \phi_{\ky}-\left(y-\Delta y_{\ky}\right)\sin\phi_{\ky}, \nonumber\\
      &\left(x-\Delta x_{\ky}\right)\sin\phi_{\ky}+\left(y-\Delta y_{\ky}\right)\cos\phi_{\ky}\bigr). \nonumber
\end{align}
Eq.~(\ref{eq:motion2}) forms a translated and rotated version of the desired image~$I$. A pure translation without rotation in the image space corresponds to a phase shift in the frequency space:
\begin{equation}
    \label{eq:motion-translation}
    \Fhat_{t}\left(k_x,k_y\right) = F\left(k_x,k_y\right)\exp\left\{-j2\pi\left(k_x\frac{\Delta x_{\ky}}{N}+k_y\frac{\Delta y_{\ky}}{N}\right)\right\} 
\end{equation}
for a~$N\times N$ image. A pure rotation about the center of the image space without translation corresponds to a rotation by the same angle in the frequency space:
\begin{equation}
    \label{eq:motion-rotation}
    \begin{aligned}
    \Fhat_{r}\left(k_x,k_y\right) = F\bigl(&k_x\cos\phi_{\ky}-k_y\sin\phi_{\ky}, \\
    &k_x\sin\phi_{\ky}+k_y\cos\phi_{\ky}\bigr).
    \end{aligned}
\end{equation}

To simulate motion artifacts during image acquisition as described in Eq.~(\ref{eq:motion2}), we sample three motion parameters at various lines in frequency space: a horizontal translation~$\Delta x$, vertical translation~$\Delta y$, and rotation~$\phi$. We report results for the case when the fraction~$\gamma_{m}$ of the total number of lines at which motion occurs is 0.03, though the trends in our results hold for several different values of this parameter. We apply the sampled motion parameters to contiguous lines in frequency space between consecutive motion line samples. Translation parameter values are drawn uniformly from the range~$[-8\textrm{px}, 8\textrm{px}]$, corresponding to physical translations on the range~$[-8\textrm{mm}, 8\textrm{mm}]$. Rotation parameter values are drawn uniformly from the range~$[-11^{\circ},11^{\circ}]$. These parameter ranges are chosen to include extreme motion at the upper limit of what might be expected in a typical MRI scan. For a Cartesian, fully-sampled acquisition, the resulting combined frequency space data represents the signal acquired when the imaging subject shifts according to the sampled motion parameters at each of the randomly sampled lines in frequency space. 

\paragraph{Noise.} \label{section:BrainNoise}
Noisy MRI data can be modeled via an additive i.i.d. complex Gaussian distribution:
\begin{equation}
\label{eq:noise}
\begin{aligned}
    \Fhat\left(\kx,\ky\right)= \F\left(\kx,\ky\right)+\epsilon_1+j\epsilon_2,\\
    \epsilon_1,\epsilon_2 \sim \mathcal{N}(0,\sigma^2 \mathbb{I}_{N\times N}),~\epsilon_1 \perp \!\!\! \perp \epsilon_2,
    \end{aligned}
\end{equation}
where~$\mathcal{N}(\mu,\Sigma)$ represents the Gaussian distribution with mean~$\mu$ and covariance~$\Sigma$. This noise distribution gives rise to the standard Rician distribution on MRI image space pixel magnitudes~\citep{cardenas2008noise}. 

To simulate noisy acquisitions as described in Eq.~(\ref{eq:noise}), we sample pixelwise independent noise from a zero-mean Gaussian distribution. We report results in the case where this noise has standard deviation~$\gamma_{n}$ of 10,000, though our observed trends are consistent for both smaller and larger values of this parameter. This value was chosen because it visually results in an aggressive noise corruption on the magnitude image; the average resulting magnitude image has SNR$\approx$1.5.

\paragraph{Undersampling. } \label{section:BrainUndersample}

To speed up image acquisition, a common approach is to only acquire data at a subset $S_y$ of discrete ``lines," i.e., values of $\ky\in S_y$:
\begin{equation}
    \label{eq:undersampling}
    \Fhat\left(\kx,\ky\right) =\begin{cases} 
      \F\left(\kx,\ky\right) & \ky\in S_y \\
      0 & \ky\not\in S_y.
   \end{cases}
\end{equation}

We simulate undersampling as described in Eq.~(\ref{eq:undersampling}) with sampling frequency~$\gamma_{s}= 25\%$ (equivalent to an acceleration factor of 4), where the selected line indices~$S_y$ are sampled at random. These lines are selected without a bias toward the low-frequency lines at the center of the Fourier plane of each image, independently of the sampling pattern in all other images. This challenging undersampling pattern measures how well different layer architectures perform under non-traditional acquisition schemes, for example, when using scan-specific acquisition patterns~\citep{bahadir2020deep}. Our subsequent experiments evaluate the proposed layers with more conventional undersampling schemes.
As an aside, the ground truth data in this experiment has conjugate symmetry in the frequency space, so in the hypothetical case of~$\gamma_{s}$=50\% with our random sampling scheme it is possible that all of the data required to perfectly reconstruct the image is present in the input. This is impossible for the acceleration factor of $\gamma_{s}$=25\% in this study. 

\begin{figure}[t]
    \centering
    \includegraphics[width=0.9\columnwidth]{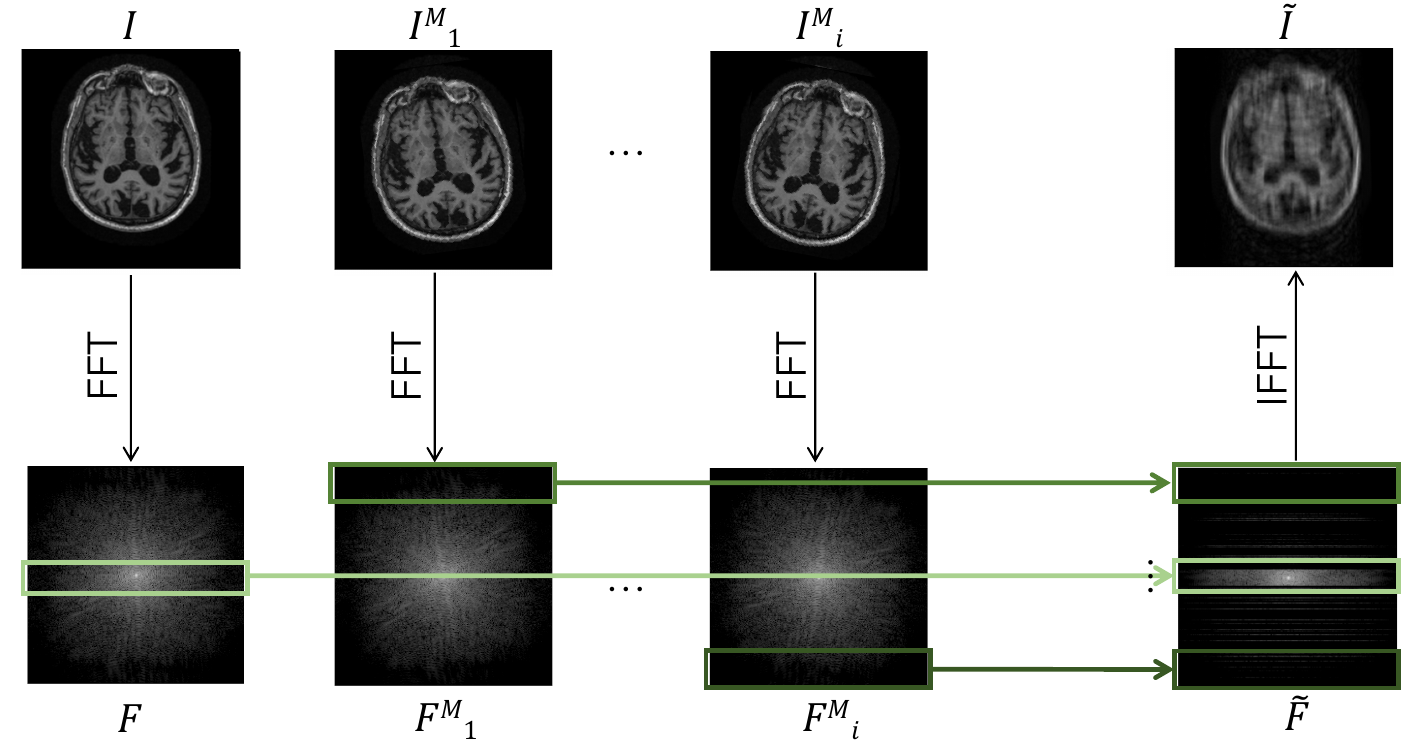}
    \caption{Data generation procedure for undersampling in the presence of motion. At line $L_i$ in frequency space, the original image $I$ is rotated and translated to form $I^M_i$. Lines from the corresponding Fourier transforms $F$ and $F^M_i$ are mixed and undersampled to generate motion-corrupted frequency space data $\tilde{F}$ that would have been acquired under the illustrated motion pattern. A similar method is used to simulate pure motion corruption without undersampling, where all frequency space lines are maintained to generate $\tilde{F}$.}
    \label{fig:motion_diagram}
\end{figure}

\paragraph{Undersampling with Motion.}\label{section:BrainUndersamplingMotion}
Undersampling reduces scan time and thus is commonly used to limit the time during which motion can occur. We analyze the setting where both motion corruption and undersampling occur simultaneously (Fig.~\ref{fig:motion_diagram}), forcing the reconstruction algorithms to correct both types of artifacts. As in the pure motion experiments, for each slice, we set the fraction of lines $\gamma_{m}=0.03$. For each line affected by motion, we sample three parameters of motion: $\Delta x_i$, $\Delta y_i$, and $\phi_i$, corresponding respectively to a horizontal translation, vertical translation, and counterclockwise rotation about the slice origin. We simulate the corresponding motion-corrupted frequency space as described in Eq.~(\ref{eq:motion2}). We then sample the full center 8\% of $k_y$-lines and sample the remainder of the line indices from a uniform distribution to achieve an overall 4x acceleration factor.

\subsubsection{Results}
\begin{figure*}
\begin{center}
\includegraphics[width=\linewidth]{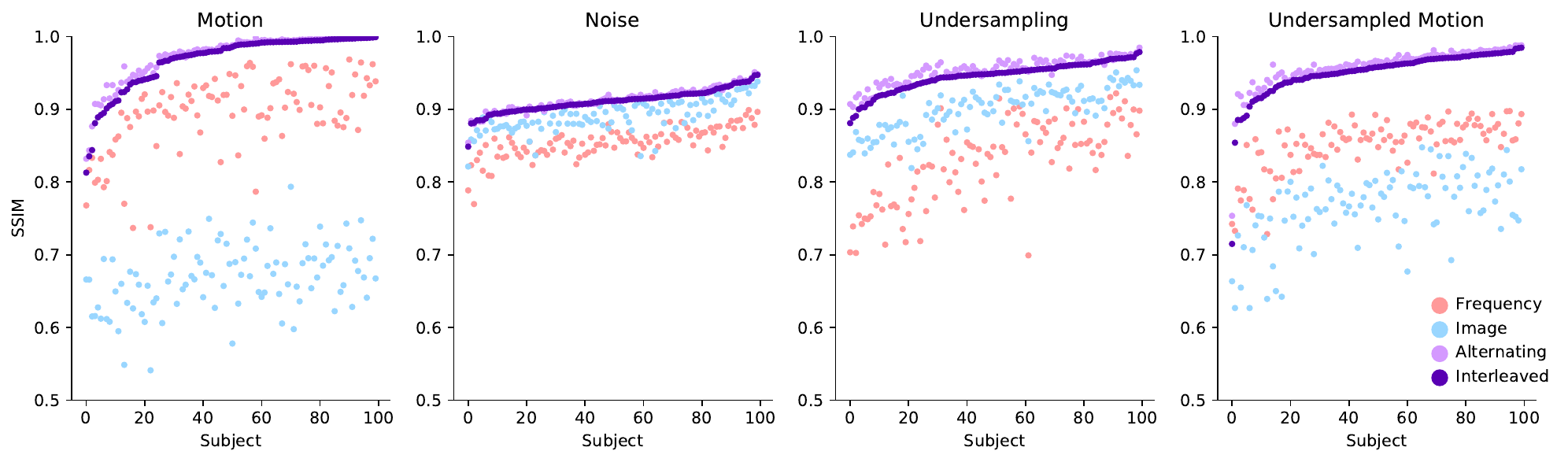}
\end{center}
\caption{Subjectwise SSIM comparison for all brain MRI tasks without data consistency constraints. Subjects are sorted by performance of the \Joint~network. For all tasks, networks combining frequency and image space convolutions outperform single-domain networks.}
\label{fig:subjectwise}
\end{figure*}

\begin{figure*}[t]
    \begin{center}
    \includegraphics[width=\linewidth]{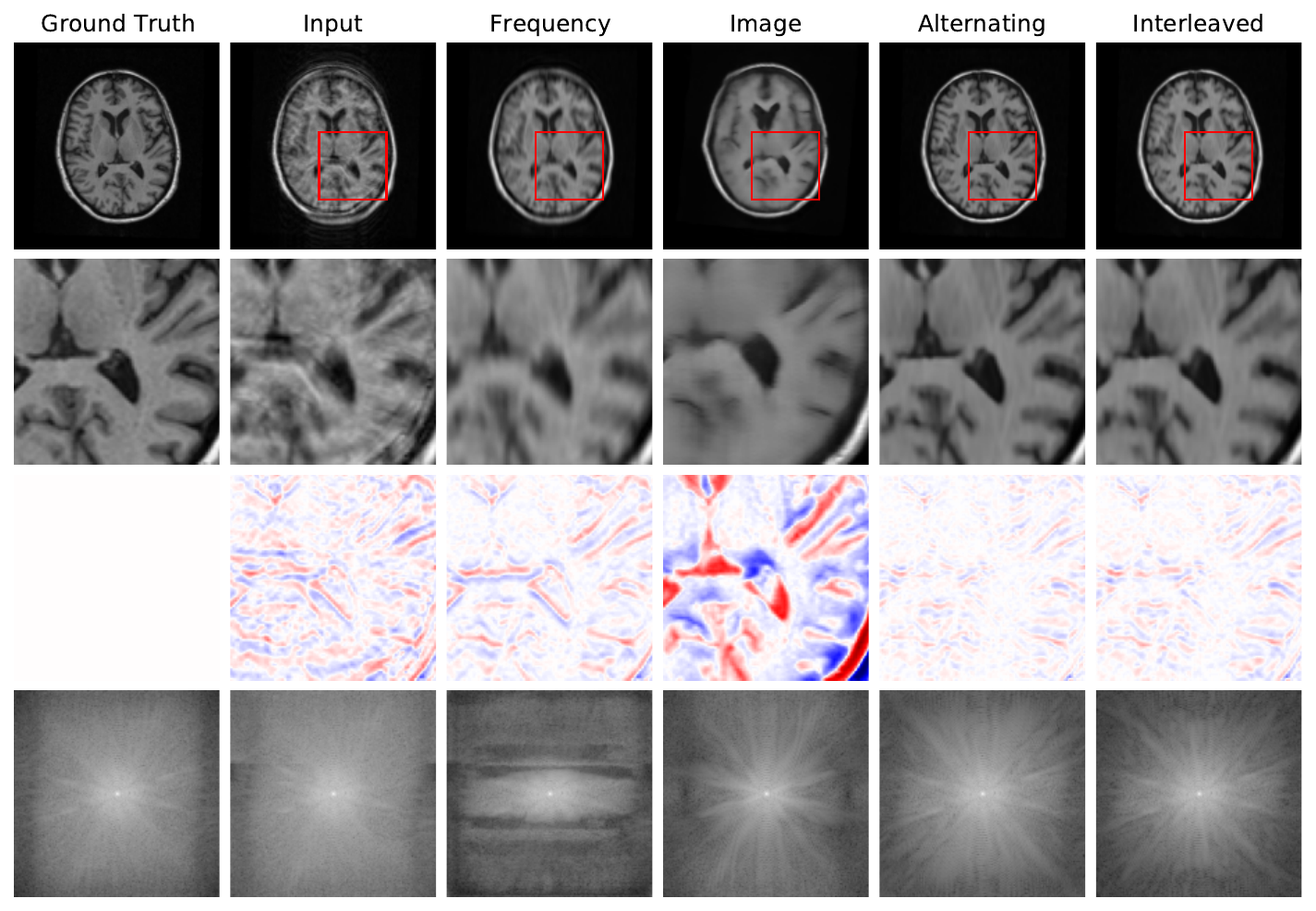}
    \label{fig:motion_img}
    \end{center}
    \vspace{-20pt}
\caption{Example reconstructions with motion at 3\% of scanning lines, zoomed-in image patches, difference patches between reconstructions and ground truth images, and frequency space reconstructions. The log values are taken of the frequency space data to better visualize its dynamic range. In the patch difference, red pixels have a higher value in the reconstruction than in the ground truth, while blue pixels have a lower value in the reconstruction than in the ground truth. The \Joint~and~\Alternating~architectures more accurately eliminate the `shadow' of the moved brain and the induced blurring compared to the single-domain networks.}
\label{fig:motion}
\end{figure*}

\begin{figure*}[t]
    \begin{center}
    \includegraphics[width=\linewidth]{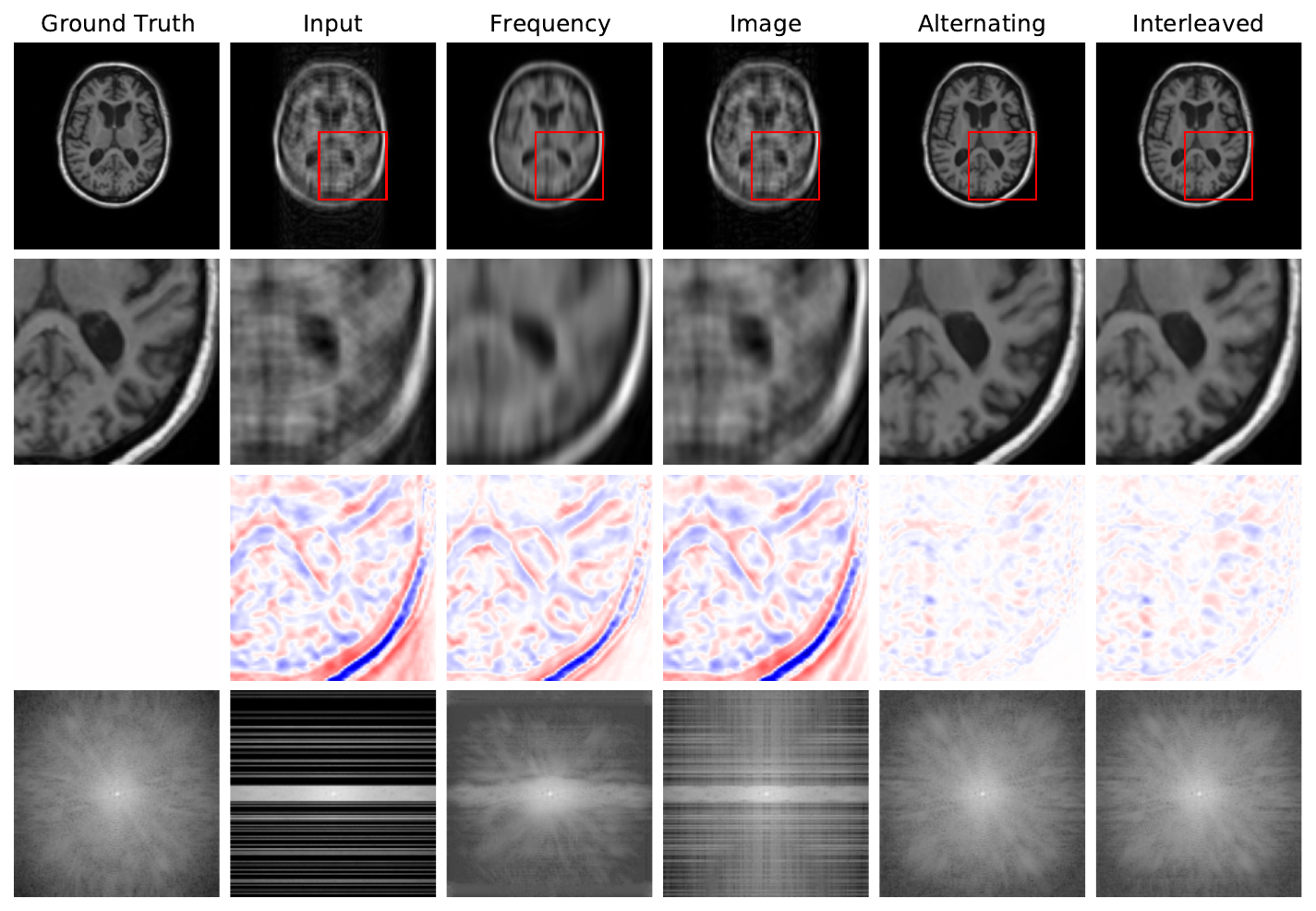}
    \label{fig:brain_undersample_motion_img}
    \end{center}
\caption{Example reconstructions from 4x undersampled, motion-corrupted data data, zoomed-in image patches, difference patches between reconstructions and ground truth images, and frequency space reconstructions. As in the motion corruption and undersampling examples, the \Joint~and~\Alternating~architectures provide more accurate reconstructions of the ground truth images and reconstructing a more coherent k-space.}
\label{fig:brain_undersample_motion}
\end{figure*}

\begin{figure*}[t]

    \includegraphics[width=\linewidth]{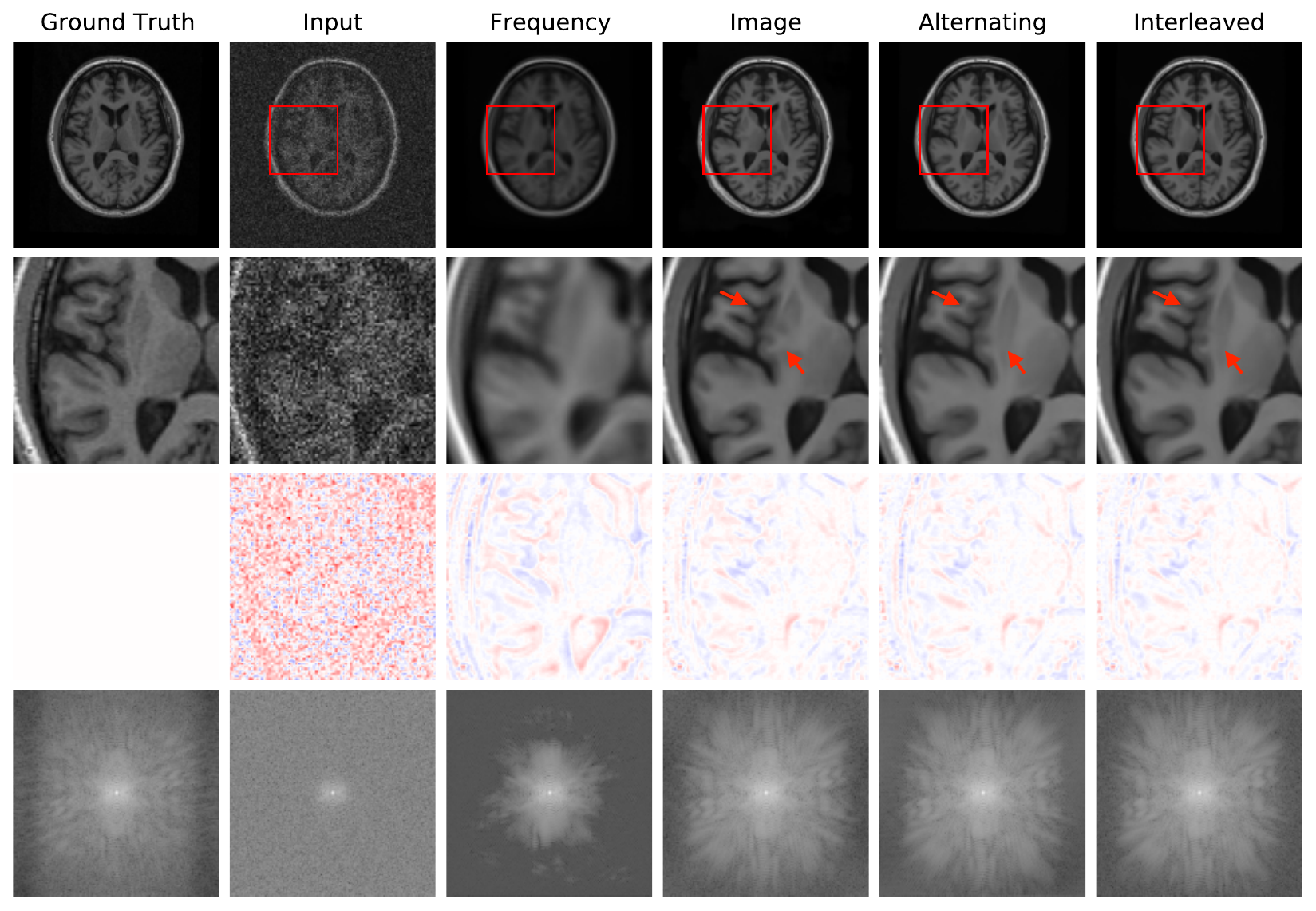}
    \label{fig:denoise_img}

\caption{Example reconstructions with noise of standard deviation 10,000. The \Joint~and~\Alternating~ reconstructions remove the pixelated noise effect without over-smoothing, in contrast to the single-domain networks.}
\label{fig:denoise}
\end{figure*}

\nw{Fig.~\ref{fig:subjectwise} reports reconstruction quality statistics for all four types of simulations described in Section~\ref{section:ExperimentalSetup}: motion, noise, undersampling, and motion combined with undersampling.} The \Joint~and \Alternating~architectures outperform the baseline architectures for nearly every task and subject. Across all tasks and nearly all subjects, the \Joint~and \Alternating~architectures are quite similar in numerical performance. Sample image reconstructions for the motion, motion with undersampling and denoising tasks are shown in Figs.~\ref{fig:motion}-\ref{fig:denoise}. Qualitatively, for each task, the \Freq~network provides a blurry version of the ground truth image. The \Image~network provides a reconstruction which effectively removes `background' effects but has limited success in correcting these artifacts within the image. In contrast, the \Joint~and \Alternating~networks provide sharper, high-quality reconstructions across all tasks. Further, the frequency space reconstructions provided by those networks appear the most faithful to the ground truth frequency data.

\subsection{Hard Data Consistency Constraint}

Deep learning for undersampled reconstruction is an active area of research and several state of the art methods have emerged for this task. In this experiment, we 
compare \Joint~and \Alternating~networks to such methods on  ADNI data introduced in Section~\ref{sec:exp1}. 

\nw{Undersampling is fundamentally different from motion and noise corruption, because the acquired data for lines $k_y\in S_y$ are the correct, desired outputs of the reconstruction algorithm at those frequency space locations. Data consistency can be enforced at test time and at intermediate layers of the network by substituting the appropriate k-space lines into the k-space representations of the image (final or intermediate) produced by the network. We enforce data consistency in \Joint~and \Alternating~networks by copying the acquired frequency space data  into the network output.}

We compare \Joint~and \Alternating~networks to a U-Net~\citep{falk2019u}, the CascadeNet~\citep{schlemper2017deep}, which combines image space convolutions with forced data consistency at each layer of the network, and, most similar to our method, the KIKI network~\citep{eo2018kiki}, which includes two separate image and frequency space networks. \nw{The KIKI-net architecture incorporates four networks operating in the frequency, image, frequency, and image spaces, respectively. This is in contrast to our networks, where every layer contains convolutions in both spaces and uses a custom nonlinearity for the frequency space layers. Moreover, the KIKI-net architecture imposes a data consistency constraint after each k-space subnetwork.  For tasks other than undersampled image reconstruction, the data consistency constraints in CascadeNet and KIKI-net would incorrectly force  the acquired k-space lines to be maintained in the final reconstruction; thus, we restrict comparisons with CascadeNet and KIKI-net to the undersampled reconstruction case.}

We use implementations of the baseline methods available at \url{https://github.com/zaccharieramzi/fastmri-reproducible-benchmark} \citep{ramzi2020benchmarking}. We scale each network to have roughly 800,000 parameters for fair comparison with our joint architectures. We use an L1 loss function to train the networks and SSIM scores to evaluate their performance on the test set.

\nw{\paragraph{Undersampling patterns. } In addition to the random sampling scheme in Section~\ref{sec:exp1}, we simulate two traditional undersampling patterns: (i) the central 8\% of lines are fully sampled while every fourth line of the outer regions of k-space is sampled and (ii) the central 4\% of lines are fully sampled while every eighth line of the outer regions of k-space is sampled.}

\subsubsection{Results}

\begin{figure*}
    \begin{center}
    \includegraphics[width=\linewidth]{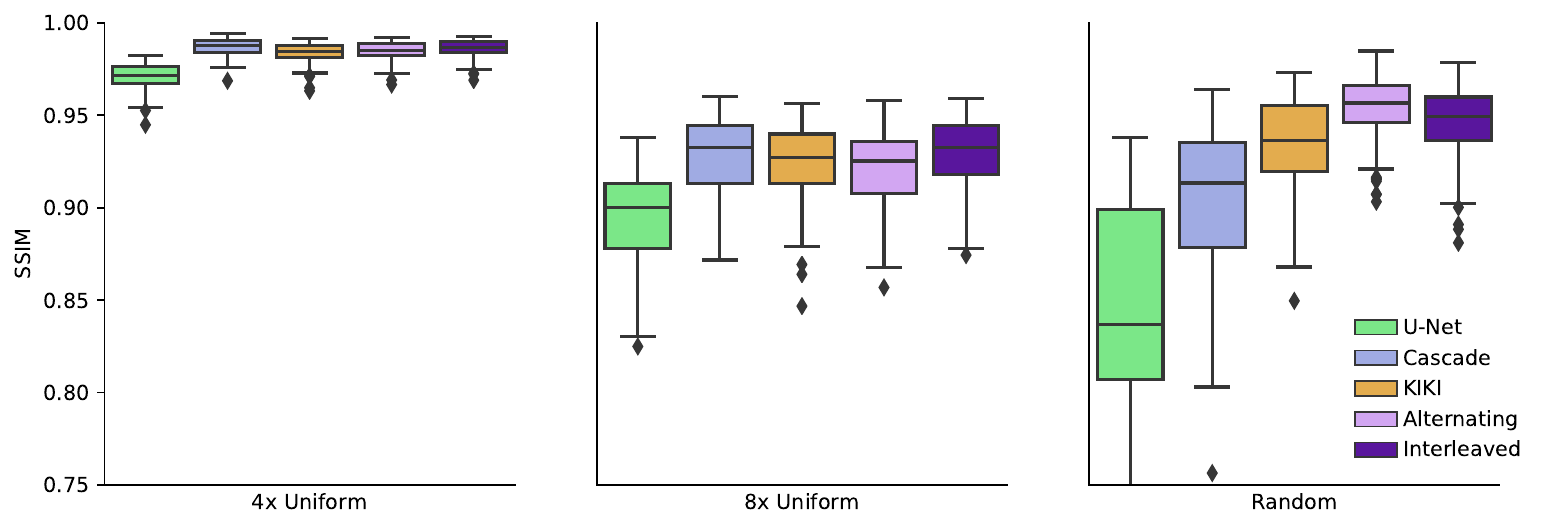}
    \end{center}
    \vspace{-10pt}
\caption{SSIM comparison of the joint networks with the state of the art undersampled reconstruction approaches on ADNI data. Results are reported for three undersampling patterns: 4x uniform undersampling with a fully-sampled central region (left), 8x uniform undersampling with a fully-sampled central region (middle), and 4x undersampling at random (right).  In all cases, simple networks composed of repeated copies of our joint layers perform at least as well as other state of the art networks, and in the difficult case of a random sampling pattern, outperform the baseline networks.}
\label{fig:undersample_scatter}
\end{figure*}

\begin{figure*}
    \begin{center}
    \includegraphics[width=\linewidth]{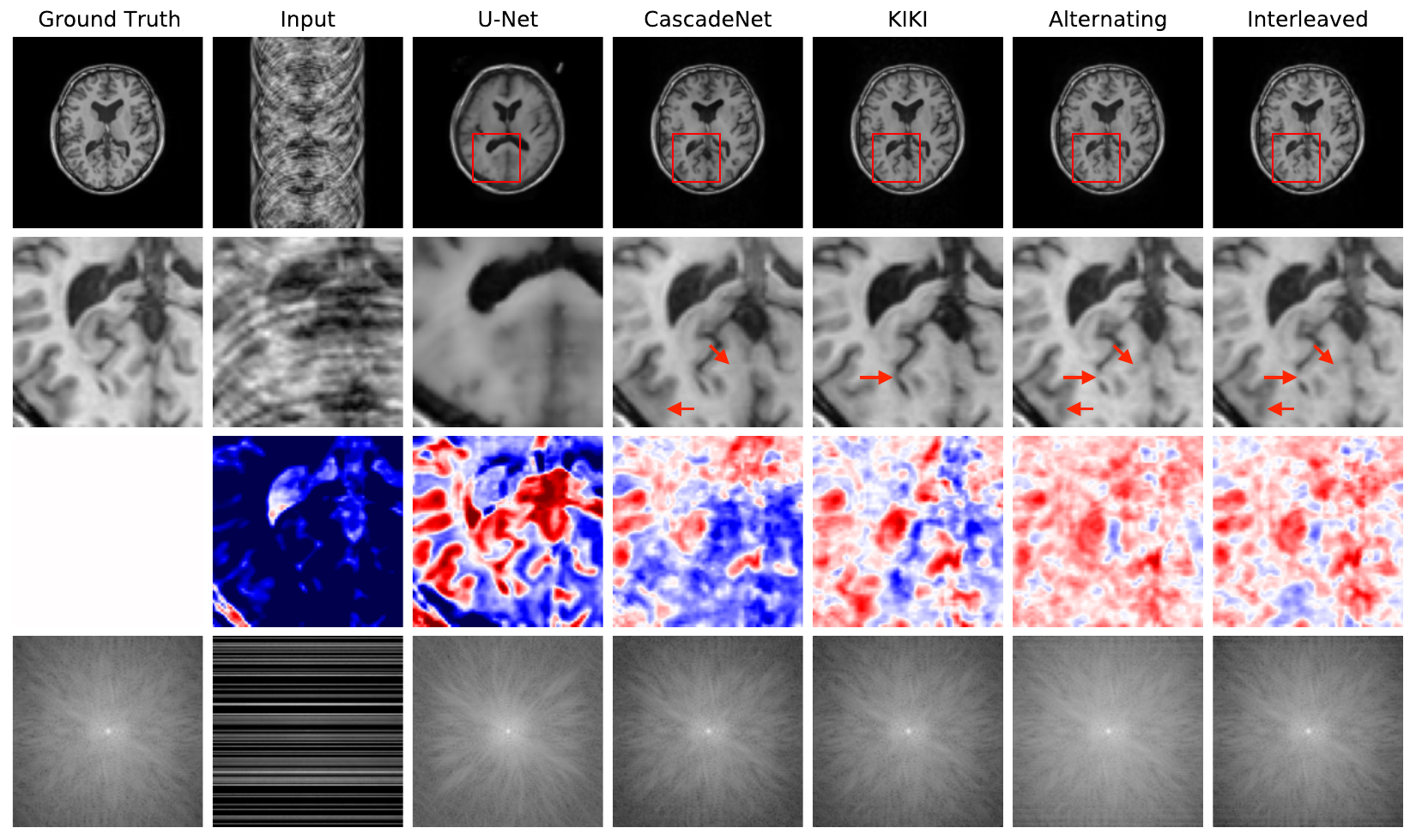}
    \label{fig:brain_undersample_img}
    \end{center}
    \vspace{-20pt}
\caption{Example reconstructions from 4x undersampled data, with lines selected at random. The \Joint~and~\Alternating~architectures provide more accurate reconstructions of the ground truth images, better eliminating `ringing' and blurring artifacts.}
\label{fig:brain_undersample}
\end{figure*}

Fig.~\ref{fig:undersample_scatter} reports statistics for U-Net, CascadeNet, KIKI-net, Joint~and \Alternating~networks.  Fig.~\ref{fig:brain_undersample} provides sample image reconstructions.  \Joint~and \Alternating~networks perform comparably to other state of the art methods on the simpler uniform undersampling tasks and outperform the state of the art methods on the more complex random undersampling task.

\subsection{Unrolled Optimization}

\nw{Finally, we  evaluate the performance of the proposed joint layers in the setting of an unrolled optimization architecture on real world multicoil MRI data. In this experiment, we replace the image space convolutional layers with our \Joint~layers in the 
MoDL framework~\citep{aggarwal2018modl} for unrolled optimization. We use the authors' publicly available implementation of MoDL at \url{https://github.com/hkaggarwal/modl}. Each iteration of the MoDL network first passes the input through convolutional layers that serve as a data-driven regularizer and then applies an analytical update based on the data consistency term.  To keep the total number of convolutions comparable, we train the baseline MoDL  network with 10 image convolutional layers in each iteration and the joint MoDL network  with 5 \Joint~layers in each iteration. We set $K=5$ iterations for both networks. The authors use the strategy of first training  a one-iteration MoDL network and using its weights to initialize the training of a multi-iteration MoDL network. This process speeds up training of the larger unrolled optimization network and avoids instabilities. We found that pre-training of a one-iteration model was unnecessary when using the joint layers, and train both the one-iteration and the five-iteration joint MoDL networks using random initializations. For consistency with the original MoDL training approach, we train all networks using L2 loss. 
}

\nw{\paragraph{Data. } We use the data from the original MoDL study~\citep{aggarwal2018modl}. This dataset contains raw k-space data from 3D T2 CUBE acquisitions with Cartesian readouts using a 12-channel head coil. The dataset contains 360 training slices from 4 training subjects and a single, separate test subject. We exclude some edge slices in this test volume and use the central 90 slices for our evaluations to match the training distribution. We train all networks using a variable density 6x undersampling mask as specified in the original paper.}

\subsubsection{Results}

\begin{figure*}[t]
    \begin{center}
    \includegraphics[width=\linewidth]{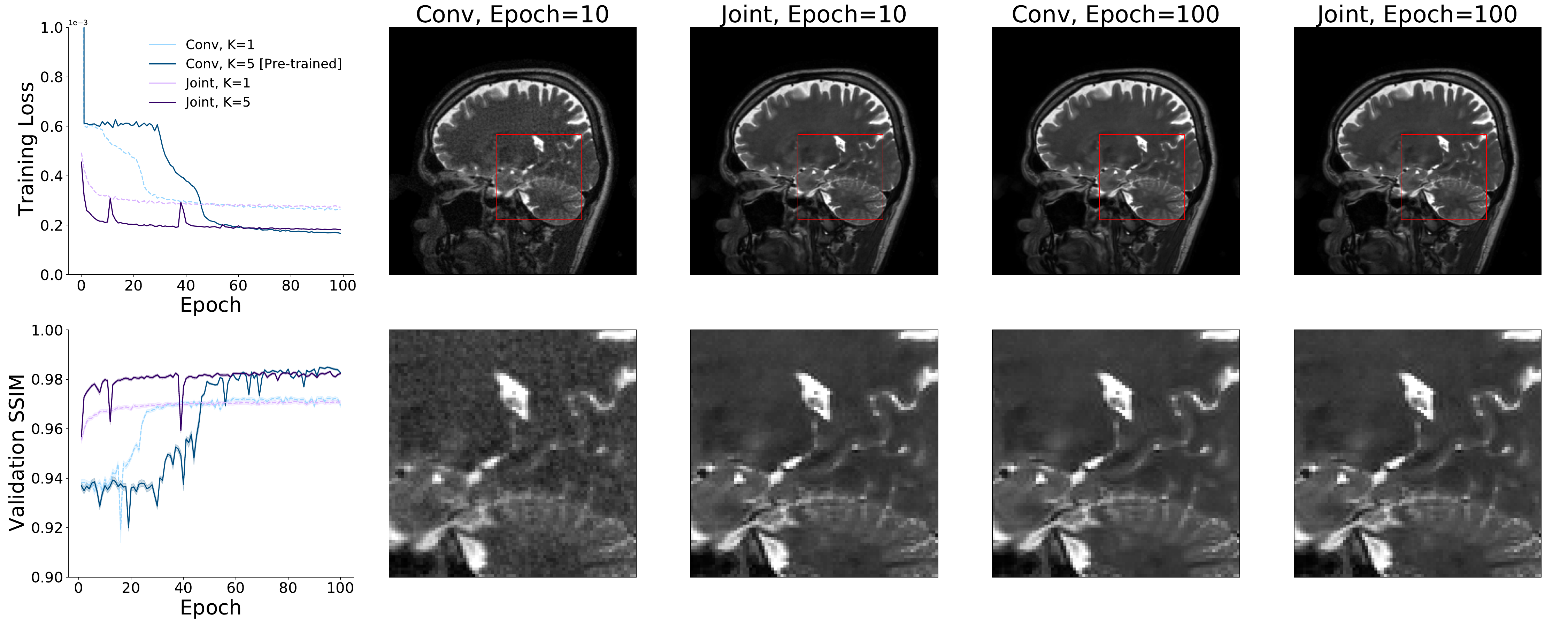}
    \end{center}
    \vspace{-10pt}
\caption{Example training loss and validation SSIM curves (left) and sample reconstructions and patches for MoDL networks with $K=1,5$ iterations trained with image convolutional layers and with the proposed joint (\Joint) layers. MoDL networks with image  convolutional layers do not converge if trained directly with $K=5$. Instead, a $K=1$ MoDL network must be trained and used to initialize the weights of a $K=5$ MoDL network. MoDL networks trained with joint layers do not require pre-training and achieve the same loss and validation SSIM values as networks trained with image convolutions in significantly less time.}
\label{fig:modl}
\end{figure*}

\nw{Figure~\ref{fig:modl} presents the training curves, validation SSIM, and sample reconstructions for all versions of the MoDL architecture. All networks attain similar validation SSIM values, but MoDL networks with joint layers achieve high reconstruction quality in roughly a third as many epochs as image space networks. Further, using our joint layers removes the need to pretrain a one-iteration network. The five-iteration network with joint layers trains successfully from random initializations. The resulting differences in wall clock training times are summarized in Table~\ref{table:modl}.
}

\begin{table}[t]
\centering
\begin{tabular}{cccc}
\hline
\textbf{MoDL Layer} & \textbf{Pre-Training (Hrs)} & \textbf{Training (Hrs)} & \textbf{Total (Hrs)} \\ \hline
Image Convolution   & 19                          & 12                      & 31                   \\ \hline
Joint Layer         & 0                           & 4                       & 4                    \\ \hline

\end{tabular}
\vspace{12pt}
\caption{\nw{Training times for the full ($K=5$) versions of the MoDL architecture to achieve validation SSIM$\,\geq 0.98$. For stable training, MoDL with image space convolutions must be initialized using the weights learned for a $K=1$ MoDL network. MoDL architectures trained with our joint layers require no pre-training. In total, using joint layers results in roughly an 8x speed-up over the pure image space approach.}}
\label{table:modl}
\end{table}

\section{Discussion and Conclusions}
We demonstrate the advantages of joint image and frequency space learning strategies for correcting corrupted MRI data. \nw{For tasks where data consistency constraints cannot be readily applied, our joint networks produce sharper reconstructions than the more blurry, artifacted versions generated by single space networks. For the well-studied task of undersampled reconstruction, where data consistency constraints can be imposed easily, we show that networks comprising joint layers can be trained with such constraints and compare favorably to other strategies that incorporate data consistency constraints to improve the quality of single space network reconstructions. For unrolled architectures that iteratively perform the steps of an optimization procedure to produce high quality reconstructions, the joint layers can straightforwardly replace image convolutional layers to improve training landscape and convergence.} These results point to joint layers as a useful building block when designing neural network architectures for correcting frequency space artifacts.

While we demonstrate our method in a diverse set of acquisition scenarios, our analysis does not exhaustively cover all possible imaging artifacts. For example, we do not analyze the effects of interslice motion, which may occur in addition to the intraslice motion studied in this work and introduces new image content from an adjacent slice into the slice being imaged. Further, while we analyze extremely aggressive versions of motion, noise, and undersampling to demonstrate the effectiveness of our method in the most challenging scenarios, future versions of this method could tune these parameters to more closely match the statistics of the patient population being scanned. For example, empirically measured motion trajectories could be used to characterize the rate and severity of the induced motion artifacts.

In the future, we aim to develop additional strategies for applications where direct consistency with acquired data is not necessarily desirable, such as motion correction. We also plan to investigate local operations beyond convolutions that more directly capitalize on properties and symmetries of frequency space data for use in joint architectures. \nw{Local convolutions in the frequency space represent a subset of all possible element-wise multiplications in the image space. Thus, future work could perform these operations in the image space, saving the computational overhead of performing an FFT within each layer, or could take advantage of additional element-wise image space multiplications whose Fourier transforms are not bandlimited to the size of our filter kernels.} The combination of these advances promises to significantly improve reconstruction and analysis of MRI data in the face of widely varying acquisition challenges and downstream applications.

\acks{The authors thank members of the Medical Vision Group at MIT CSAIL for useful discussions. This research was supported by NIBIB, NICHD, NIA, and NINDS of the National Institutes of Health under award numbers 5T32EB1680, P41EB015902, R01EB017337, R01HD100009, R01EB032708, 1R01AG064027-01A1, 5R01NS105820-02, 1R01AG070988-01 and 1RF1MH123195-01, by the European Research Council Starting Grant 677697, project “BUNGEE-TOOLS," by Alzheimer's Research UK (ARUK-IRG2019A-003), by an NSF Graduate Research Fellowship, and by a Google PhD Fellowship.}

\ethics{The work follows appropriate ethical standards in conducting research and writing the manuscript, following all applicable laws and regulations regarding treatment of animals or human subjects.}

\coi{We declare we don't have conflicts of interest.}

\bibliography{fourier}

\clearpage
\onecolumn

\appendix

\section{FastMRI Experiments}

We compare the \Joint~and \Alternating~networks with \Freq~and \Image \ baseline methods, as well as the top three methods submitted to the single coil track of the FastMRI challenge at \url{https://fastmri.org}.

\subsection{Data} We train and evaluate all networks on the proton density knee MRI frequency space data from the single coil FastMRI Dataset~\citep{zbontar2018fastmri}. We train separate networks for signals acquired with and without fat suppression.
We apply the FastMRI 4x undersampling scheme at both training and test time. The 4x undersampling scheme acquires all of the central 8\% of lines and samples lines outside of the central region from a uniform distribution such that 25\% of all lines are sampled in total. After undersampling the signals, we normalize each input and output training pair by dividing by the maximum value in the corrupted image. We use the standard FastMRI split of 34,742 training slices from 973 volumes and 7,135 validation slices from 199 volumes. No subjects are shared across these sets. We treat the FastMRI validation set as our test set and use it only for evaluation by comparing the network's output to the high quality fully sampled images provided as part of the FastMRI dataset.

\subsection{Training Loss and Evaluation Metrics}
We evaluate and compare the networks trained with a variety of loss functions and assess reconstruction quality via different  quality metrics. \nw{We train \Freq, \Image, \Joint~and \Alternating~networks with seven loss functions: image space L1 error, frequency space L1 error, a joint L1 metric summing image and frequency L1 errors, SSIM~\citep{wang2004image}, multiscale SSIM~\citep{wang2003multiscale}, and PSNR~\citep{huynh2008scope}.} The joint L1 metric weighs the frequency space L1 error by 0.1 relative to the image space L1 error to account for differences in the error magnitudes. The SSIM and multiscale SSIM scores are computed with window size~$7 \times 7$ and constants~$k_1 = 0.01$,~$k_2 = 0.03$.

We also compare the joint networks with top single coil methods on the FastMRI benchmark. For these experiments, we use a larger version of the~\Joint \ network comprised of 6 joint layers with two frequency space and two image space convolutions per layer, yielding roughly 3 million parameters total.

\subsection{Results}
\begin{figure*}[h]
    \centering
    \includegraphics[width=\textwidth]{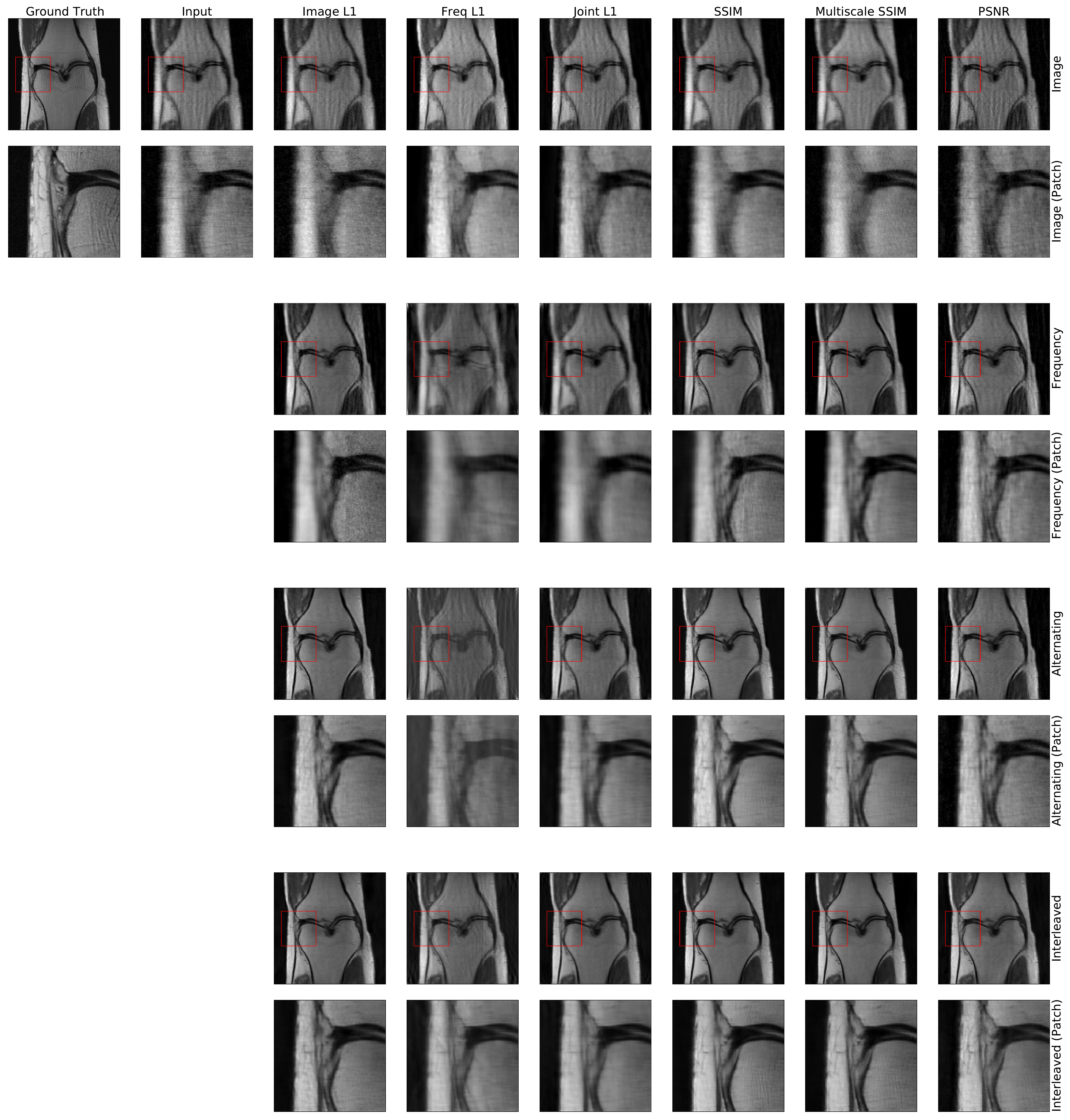}

    \caption{Typical image reconstruction results for all architectures (rows) and loss functions (columns) on FastMRI images without fat suppression. The \Joint~and \Alternating~networks provide the sharpest reconstructions for all loss functions. Amongst these, both SSIM-based loss functions most sharply reconstruct high frequency structures within the zoomed-in patch. Similar results are observed in images with fat suppression.}
    \label{fig:knee_loss_expanded}
\end{figure*}

Our results on the knee undersampled reconstruction task replicate the trends observed in the brain undersampled reconstruction task. Joint networks outperform single-domain networks, as reported in Table~\ref{table:loss_table}. This suggests that our joint layers can successfully process acquired, complex-valued MRI data. Further, Table~\ref{table:loss_table} confirms that the success of  joint learning is not specific to a certain loss landscape. Qualitative examples of reconstructions from networks trained with various loss functions are shown in Fig.~\ref{fig:knee_loss_expanded}.

\begin{figure*}
    \centering
    \includegraphics[width=\textwidth]{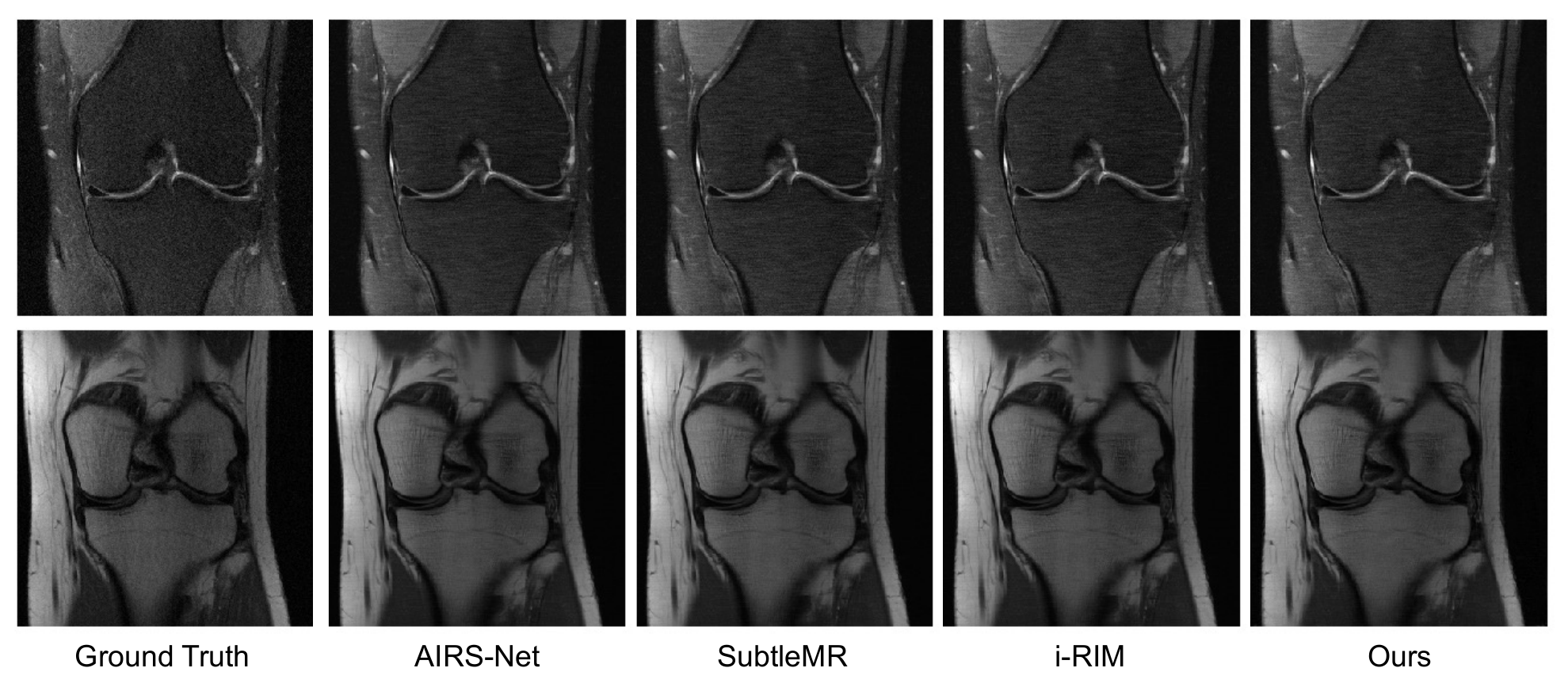}

    \caption{Comparison of the \Joint~reconstruction results with the top methods on the FastMRI single coil knee reconstruction challenge. All images were taken from the FastMRI online submission website. Our method produces a reconstruction qualitatively similar to those of the top three  methods on the leaderboard.}
    \label{fig:knee_sota_comparison}
\end{figure*}

\nw{The reconstructed images produced by the larger \Joint~network are qualitatively similar to those produced by the top three methods on the FastMRI leaderboard (Fig.~\ref{fig:knee_sota_comparison}).} Table~\ref{table:fastmri_comparison} reports reconstruction quality measures for  \Joint~network and the top single-slice methods on the FastMRI benchmark.  \Joint~network achieves results that are close to the state of the art architectures specifically tuned for this task. We emphasize that our goal is not to attain state of the art performance on the FastMRI benchmark, but rather to show that simple layers comprised of both frequency and image space convolutions achieve reasonable performance on this benchmark while offering flexibility for correcting a wide range  of other artifacts, and for correcting multiple artifacts present simultaneously.

\begin{landscape}
\begin{table}
\centering

\begin{tabularx}{\linewidth}{cccccccc} \hline \textbf{Loss} & \textbf{Architecture} &\thead{\textbf{Freq L1 ($\downarrow$)}} &\thead{\textbf{Image L1 ($\downarrow$)}} &\thead{\textbf{Joint L1 ($\downarrow$)}} &\thead{\textbf{SSIM ($\uparrow$)}} &\thead{\textbf{MS SSIM ($\uparrow$)}} &\thead{\textbf{PSNR ($\uparrow$)}}\\ \hline 
Freq L1 & \makecell{Frequency \\ Image \\ Interleaved \\ Alternating} &\makecell{~5.2 $\pm$ 1.4 \\ ~8.9 $\pm$ 3.4 \\ ~\textbf{3.9 $\pm$ 1.5} \\ ~4.1 $\pm$ 1.5} & \makecell{~0.089 $\pm$ 0.063 \\ ~0.079 $\pm$ 0.060 \\ ~\textbf{0.040 $\pm$ 0.018} \\ ~0.095 $\pm$ 0.023} & \makecell{~0.61 $\pm$ 0.19 \\ ~0.97 $\pm$ 0.40 \\ ~\textbf{0.43 $\pm$ 0.16} \\ ~0.51 $\pm$ 0.16} & \makecell{~0.60 $\pm$ 0.12 \\ ~0.70 $\pm$ 0.14 \\ ~\textbf{0.73 $\pm$ 0.11} \\ ~0.58 $\pm$ 0.10} & \makecell{~0.76 $\pm$ 0.06 \\ ~0.88 $\pm$ 0.06 \\ ~\textbf{0.91 $\pm$ 0.05} \\ ~0.77 $\pm$ 0.09} & \makecell{19.8 $\pm$ 2.4 \\ 22.4 $\pm$ 3.3 \\ \textbf{26.6 $\pm$ 2.4} \\ 20.3 $\pm$ 1.9} \cr \hline 
Image L1 & \makecell{Frequency \\ Image \\ Interleaved \\ Alternating} &\makecell{~7.9 $\pm$ 2.5 \\ 21.9 $\pm$ 8.2 \\ ~\textbf{6.9 $\pm$ 2.7} \\ ~7.5 $\pm$ 2.9} & \makecell{~0.040 $\pm$ 0.015 \\ ~0.054 $\pm$ 0.034 \\ ~\textbf{0.031 $\pm$ 0.018} \\ ~0.032 $\pm$ 0.013} & \makecell{~0.83 $\pm$ 0.27 \\ ~2.24 $\pm$ 0.85 \\ ~\textbf{0.72 $\pm$ 0.28} \\ ~0.78 $\pm$ 0.31} & \makecell{~0.69 $\pm$ 0.12 \\ ~0.59 $\pm$ 0.14 \\ ~\textbf{0.78 $\pm$ 0.12} \\ ~0.76 $\pm$ 0.12} & \makecell{~0.88 $\pm$ 0.06 \\ ~0.85 $\pm$ 0.09 \\ ~\textbf{0.92 $\pm$ 0.06} \\ ~0.91 $\pm$ 0.06} & \makecell{26.8 $\pm$ 2.2 \\ 24.9 $\pm$ 2.7 \\ \textbf{28.9 $\pm$ 2.5} \\ 28.5 $\pm$ 2.4} \cr \hline 
Joint L1 & \makecell{Frequency \\ Image \\ Interleaved \\ Alternating} &\makecell{~5.2 $\pm$ 1.7 \\ ~8.9 $\pm$ 3.4 \\ ~\textbf{3.9 $\pm$ 1.5} \\ ~4.1 $\pm$ 1.5} & \makecell{~0.062 $\pm$ 0.068 \\ ~0.055 $\pm$ 0.060 \\ ~\textbf{0.032 $\pm$ 0.019} \\ ~0.035 $\pm$ 0.020} & \makecell{~0.58 $\pm$ 0.23 \\ ~0.95 $\pm$ 0.39 \\ ~\textbf{0.43 $\pm$ 0.17} \\ ~0.44 $\pm$ 0.17} & \makecell{~0.66 $\pm$ 0.15 \\ ~0.70 $\pm$ 0.14 \\ ~\textbf{0.77 $\pm$ 0.12} \\ ~0.75 $\pm$ 0.12} & \makecell{~0.86 $\pm$ 0.06 \\ ~0.88 $\pm$ 0.06 \\ ~\textbf{0.92 $\pm$ 0.05} \\ ~0.91 $\pm$ 0.05} & \makecell{23.1 $\pm$ 3.1 \\ 25.4 $\pm$ 3.4 \\ \textbf{27.8 $\pm$ 2.4} \\ 26.8 $\pm$ 2.5} \cr \hline 
-SSIM & \makecell{Frequency \\ Image \\ Interleaved \\ Alternating} &\makecell{~7.3 $\pm$ 1.8 \\ 12.0 $\pm$ 4.3 \\ ~\textbf{6.4 $\pm$ 2.4} \\ ~7.4 $\pm$ 2.8} & \makecell{~0.039 $\pm$ 0.018 \\ ~0.058 $\pm$ 0.059 \\ ~\textbf{0.029 $\pm$ 0.015} \\ ~0.031 $\pm$ 0.012} & \makecell{~0.76 $\pm$ 0.20 \\ ~1.25 $\pm$ 0.49 \\ ~\textbf{0.67 $\pm$ 0.25} \\ ~0.77 $\pm$ 0.29} & \makecell{~0.73 $\pm$ 0.12 \\ ~0.69 $\pm$ 0.14 \\ ~\textbf{0.80 $\pm$ 0.12} \\ ~0.79 $\pm$ 0.13} & \makecell{~0.90 $\pm$ 0.05 \\ ~0.87 $\pm$ 0.07 \\ ~\textbf{0.94 $\pm$ 0.05} \\ ~0.93 $\pm$ 0.06} & \makecell{26.5 $\pm$ 2.4 \\ 24.7 $\pm$ 3.3 \\ \textbf{29.0 $\pm$ 2.3} \\ 27.7 $\pm$ 2.1} \cr \hline 
-MS SSIM & \makecell{Frequency \\ Image \\ Interleaved \\ Alternating} &\makecell{~9.3 $\pm$ 1.5 \\ 15.6 $\pm$ 7.7 \\ ~\textbf{8.6 $\pm$ 1.8} \\ 15.0 $\pm$ 4.3} & \makecell{~0.043 $\pm$ 0.023 \\ ~0.061 $\pm$ 0.045 \\ ~\textbf{0.030 $\pm$ 0.017} \\ ~0.031 $\pm$ 0.014} & \makecell{~0.98 $\pm$ 0.16 \\ ~1.63 $\pm$ 0.81 \\ ~\textbf{0.89 $\pm$ 0.19} \\ ~1.54 $\pm$ 0.44} & \makecell{~0.69 $\pm$ 0.14 \\ ~0.61 $\pm$ 0.13 \\ ~\textbf{0.79 $\pm$ 0.12} \\ ~\textbf{0.79 $\pm$ 0.12}} & \makecell{~0.91 $\pm$ 0.05 \\ ~0.86 $\pm$ 0.07 \\ ~\textbf{0.94 $\pm$ 0.05} \\ ~\textbf{0.94 $\pm$ 0.05}} & \makecell{25.3 $\pm$ 2.0 \\ 24.0 $\pm$ 3.5 \\ \textbf{27.5 $\pm$ 1.9} \\ 23.8 $\pm$ 1.8} \cr \hline 
-PSNR & \makecell{Frequency \\ Image \\ Interleaved \\ Alternating} &\makecell{~8.0 $\pm$ 2.5 \\ 15.3 $\pm$ 7.1 \\ ~\textbf{7.3 $\pm$ 2.8} \\ ~9.1 $\pm$ 4.7} & \makecell{~0.038 $\pm$ 0.012 \\ ~0.058 $\pm$ 0.043 \\ ~\textbf{0.031 $\pm$ 0.012} \\ ~0.037 $\pm$ 0.016} & \makecell{~0.84 $\pm$ 0.26 \\ ~1.59 $\pm$ 0.75 \\ ~\textbf{0.76 $\pm$ 0.29} \\ ~0.95 $\pm$ 0.49} & \makecell{~0.70 $\pm$ 0.12 \\ ~0.64 $\pm$ 0.13 \\ ~\textbf{0.77 $\pm$ 0.12} \\ ~0.70 $\pm$ 0.13} & \makecell{~0.89 $\pm$ 0.06 \\ ~0.85 $\pm$ 0.07 \\ ~\textbf{0.92 $\pm$ 0.06} \\ ~0.89 $\pm$ 0.08} & \makecell{27.4 $\pm$ 2.2 \\ 24.0 $\pm$ 2.6 \\ \textbf{29.1 $\pm$ 2.2} \\ 27.6 $\pm$ 2.4} \cr \hline 
\end{tabularx}

\vspace{4pt}
\caption{Image reconstruction evaluation metrics (columns) for networks trained with varying loss functions (rows) on images acquired without fat suppression. Similar trends hold for images with fat suppression. MS SSIM stands for multiscale SSIM. For metrics labeled $\downarrow$, smaller values are better; for metrics labeled $\uparrow$, larger values are better. Across nearly every training loss function and metric, the \Joint~network performs best. In almost every case, the \Alternating~network architecture performs similarly or only slightly worse than the \Joint~network. This is particularly true in the case of SSIM-based loss functions, which provide the best overall quantitative results across all evaluation metrics.}
\label{table:loss_table}

\end{table}
\end{landscape}

\begin{table}
\centering
\begin{tabular}{cccc}
\Xhline{2\arrayrulewidth}
\textbf{Method}    & \textbf{MAE}    & \textbf{SSIM}   & \textbf{PSNR} \\ \hline
Interleaved (Ours) & 0.0296 & 0.768 & 32.9 \\ 
AIRS-Net           & 0.0266 & 0.784 & 33.8 \\ 
SubtleMR           & 0.0270 & 0.781 & 33.7 \\ 
i-RIM              & 0.0271 & 0.781 & 33.7 \\ \hline
\end{tabular}
\vspace{4pt}
\caption{Reconstruction quality statistics on the FastMRI leaderboard test dataset, at 4x undersampling. The FastMRI dataset contains images both with and without fat suppression. Simple \Joint~network comprised of joint layers is comparable to the three top models on the FastMRI leaderboard, yielding reconstructions with SSIM within 3\% of the leading methods.}
\label{table:fastmri_comparison}
\end{table}

\end{document}